\documentclass[journal]{IEEEtran}

\usepackage{epsfig}
\usepackage{graphicx}
\usepackage{caption}
\usepackage{subcaption}
\usepackage{amsmath}
\usepackage{amssymb}
\usepackage{comment}
\usepackage{color}

\usepackage{multirow}
\usepackage{multicol}
\usepackage{rotating}
\usepackage{textcomp}
\usepackage{gensymb}
\usepackage{dsfont}
\usepackage{url}

\usepackage{algorithm2e}
\let\savedalgorithm\algorithm
\let\savedendalgorithm\endalgorithm

\usepackage{algorithm}

\begin{document}
\newcommand{\st}{{\rm s.t.}\xspace}
\newcommand{\trace}{{\bf Tr}}

\def\argmax{\operatorname*{argmax\,}}

\def\psd{\succcurlyeq}
\def\nsd{\preccurlyeq}
\def\pd{ \succ }
\def\nd{ \prec }

\def\T{{\!\top}}
\def\transpose{{\!\top}}

\def\Real{\mathbb{R}}

\newcommand{\fnorm}[2][2]{\ensuremath{ \left\| #2 \right\|_{ \mathrm{#1} } } }

\newcommand{\iprod}[1]{\ensuremath{ \left< #1 \right> } }

\def\x{{\boldsymbol x}}
\def\w{{\boldsymbol w}}

\newtheorem{theorem}{Theorem}[section]
\newtheorem{corollary}{Corollary}[section]
\newtheorem{definition}{Definition}[section]
\newtheorem{proposition}{Proposition}[section]
\newtheorem{lemma}{Lemma}[section]

\newtheorem{result}{Claim}[section]

\def\multiboost{{\sc MultiBoost}\xspace}
\def\yi{{  y_i  }}
\newcommand{\mbh}{ {\sc Multi\-Boost}-\-hinge\xspace}
\newcommand{\mbe}{ {\sc Multi\-Boost}-\-exp\xspace}

\def\brho{ {\boldsymbol{\rho} } }
\def\bbeta{ {\boldsymbol{\beta} } }
\def\bgamma{ {\boldsymbol{\gamma} } }
\def\bxi{ {\boldsymbol{\xi} } }
\def\bomega{ {\boldsymbol{\omega} } }
\def\bmu{{\boldsymbol{\mu} } }
\def\bone{ {\bf 1 } }
\def\bzero{ {\bf 0 } }
\def\ba{ {\bf a } }
\def\be{ {\bf e } }
\def\bx{ {\bf x } }
\def\bp{ {\bf p } }
\def\bq{ {\bf q } }
\def\bu{ {\bf u } }
\def\bm{ {\bf M } }
\def\bz{ {\bf z } }

\def\by{ {\bf y } }

\def\bv{ {\bf v } }
\def\bb{ {\bf b } }
\def\bw{ {\bf w } }
\def\bs{ {\bf s } }

\def\bc{ {\bf c } }

\def\bX{ {\bf X } }
\def\bS{ {\bf S } }
\def\bI{ {\bf I } }
\def\bA{ {\bf A } }
\def\bH{ {\bf H } }
\def\bW{ {\bf W } }
\def\bD{ {\bf D } }
\def\bP{ {\bf P } }
\def\bK{ {\bf K } }
\def\bL{ {\bf L } }
\def\bO{ {\bf O } }
\def\bC{ {\bf C } }
\def\bZ{ {\bf Z } }
\def\bR{ {\bf R } }
\def\bY{ {\bf Y } }

\def\bU{ {\bf U } }

\def\calX{ {\cal X } }
\def\calF{ {\cal F } }
\def\calH{ {\cal H } }
\def\calW{ {\cal W } }

\def\st{ {\rm s.t. } }
\def\eg{{\rm e.g. }}
\def\ie{{\rm i.e. }}

\def\etc{{\rm etc.}}
\def\vs{{\rm vs.}}
\def\wrt{{\rm w.r.t.}}

\def\fsv{ f_{\rm sv} }
\def\flp{ f_{\rm lp} }

\def\spx{ {\rm spx }}

\def\liblinear{{\sc liblinear}\xspace}

\def\CGENS{{\sc  CGEns}\xspace}

\def\lssvm{{\sc  CGEns-SLS}\xspace}

\def\half{{\tfrac{1}{2} }}
\def\ones{{\boldsymbol  1}}

\def\balpha{{\boldsymbol  \alpha}}

\def\bXi{{\boldsymbol  \Xi}}

\def\wl{{\hslash  }}

\def\H{{\bf H}}
\def\zeros{{\boldsymbol  0}}
\def\bh{{\bf h}}

\def\sign{{\rm sign}}

\title{Fast detection of multiple objects in traffic scenes with a common detection framework}

\author{Qichang Hu$ ^{1,2}$, Sakrapee Paisitkriangkrai$ ^{1}$, Chunhua Shen$ ^{1,3}$, Anton van den Hengel$ ^{1, 3}$,
Fatih Porikli$ ^{2, 3}$
\\
$ ^1 $University of Adelaide ~
$ ^2 $NICTA ~
$ ^3 $Australian Centre for Robotic Vision
}

\maketitle

\begin{abstract}

Traffic scene perception (TSP) aims to real-time extract accurate on-road environment information, which involves three phases: detection of objects of interest, recognition of detected objects, and tracking of objects in motion. Since recognition and tracking often rely on the results from detection, the ability to detect objects of interest effectively plays a crucial role in TSP. In this paper, we focus on three important classes of objects: traffic signs, cars, and cyclists. We propose to detect {\em all the three} important objects in a single learning based detection framework. The proposed framework consists of a dense feature extractor and detectors of three important classes. Once the dense features have been extracted, these features are shared with all detectors. The advantage of using one common framework is that the detection speed is much faster, since all dense features need only to be evaluated once in the testing phase. In contrast, most previous works have designed specific detectors using different features for each of these objects. To enhance the feature robustness to noises and image deformations, we introduce spatially pooled features as a part of aggregated channel features. In order to further improve the generalization performance, we propose an object subcategorization method as a means of capturing intra-class variation of objects.

We experimentally demonstrate the effectiveness and efficiency of the proposed framework in three detection applications: traffic sign detection, car detection, and cyclist detection. The proposed framework achieves the competitive performance with state-of-the-art approaches on several benchmark datasets.

\end{abstract}


\section{Introduction}
\label{sec:Introduction}

\begin{figure}[tbp]
\begin{center}
	\includegraphics[width=\columnwidth]{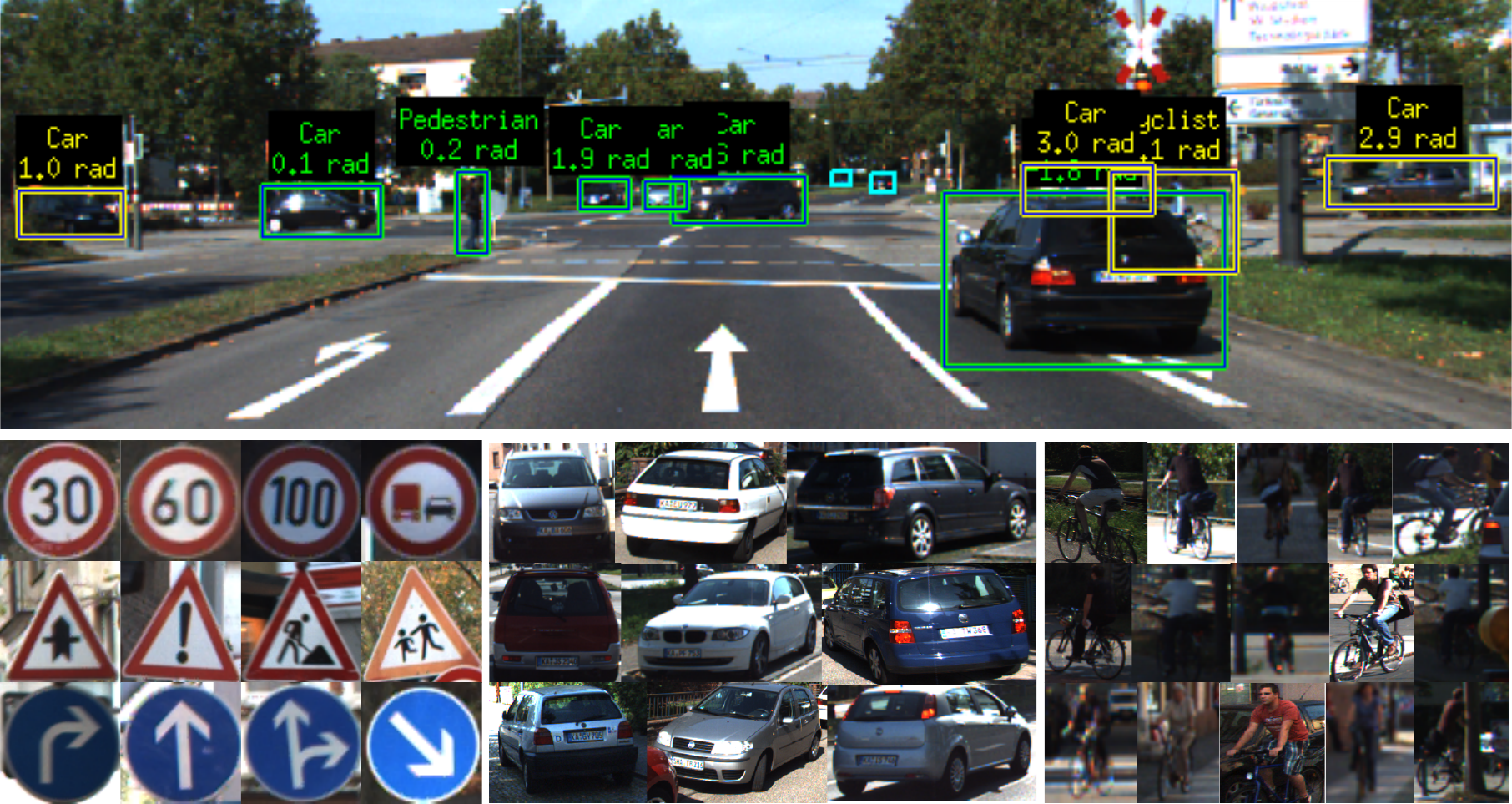}
\end{center}
\caption{Top image: A typical on-road traffic scene with the detected objects of interest. Bottom images: Each block represents one class of objects of interest. From left to right, the first block contains traffic sign examples, the second contains car examples, and the third contains cyclist examples.}
\label{fig:Introduction}
\end{figure}

Vision-based traffic scene perception (TSP) is one of many fast-emerging areas in the intelligent transportation system. This field of research has been actively studied over the past decade~\cite{sivaraman2003looking}. TSP involves three phases: detection, recognition and tracking of various objects of interest. Since recognition and tracking often rely on the results from detection, the ability to detect objects of interest effectively plays a crucial role in TSP. In this paper, we focus on three important classes of objects: traffic signs, cars, and cyclists. Fig.~\ref{fig:Introduction} shows a typical on-road traffic scene with the detected objects of interest and illustrates some positive examples from the three mentioned classes.

The aim of traffic sign detection is to alert the driver of the changed traffic conditions. The task is to accurately localize and recognize road signs in various traffic environments. Prior approaches~\cite{de1997road,de2003traffic,kuo2007two} use color and shape information. However, these approaches are not adaptive under severe weather and lighting conditions. Additionally, appearance of traffic signs can physically change over time, due to the weather and damage caused by accidents. Instead of using color and shape features, most recent approaches~\cite{mathias2013traffic,wang2013robust} employ texture or gradient features, such as local binary patterns (LBP)~\cite{ahonen2004face} and histogram of oriented gradients (HOG)~\cite{dalal2005histograms}. These features are partially invariant to image distortion and illumination change, but they are still unable to handle severe deformations.

\begin{figure}[tbp]
\begin{center}
	\includegraphics[width=0.95\columnwidth]{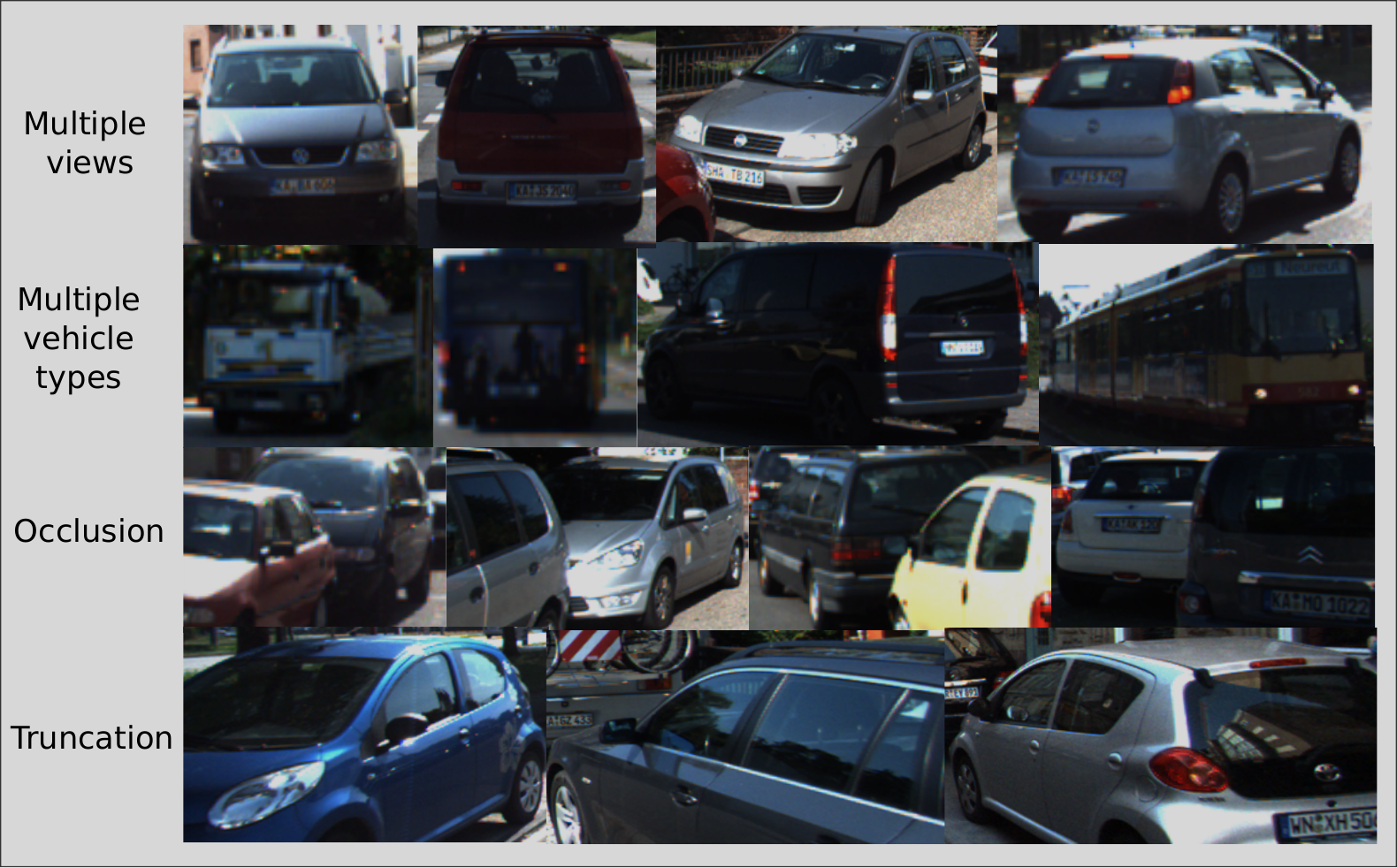}
\end{center}
\caption{A set of different vehicles with different viewpoints, occlusions, and truncations.}
\label{fig:Cars challenges}
\end{figure}

Car detection is a more challenging problem compared to traffic sign detection due to its large intra-class variation caused by different viewpoints and occlusions. Fig.~\ref{fig:Cars challenges} shows a set of different cars with a large intra-class variation. Although sliding-window based detection methods have shown promising results in face and human detection~\cite{viola2004robust,dalal2005histograms}, they often fail to detect cars due to a large variation of viewpoints. Recently the deformable parts model (DPM)~\cite{felzenszwalb2010object}, which has gained a lot of attention in generic object detection, has been adapted successfully for car detection~\cite{geiger2011joint,hejrati2012analyzing,pepikj2013occlusion}. In addition to the DPM, visual subcategorization based approaches~\cite{divvala2012important,kuo2009robust,ohn2014fast} have been applied to improve the generalization performance.

Cyclist detection is a new attractive application in the domain of TSP. At present, only few methods are designed purposely for cyclist detection. Many existing pedestrian detection approaches~\cite{dalal2005histograms,DollarPAMI14pyramids,geiger2011joint} can be adapted for cyclist detection because appearances of pedestrians are very similar to appearances of cyclists along the road. Compared to pedestrian detection, the new  problem is more difficult because the various appearances and viewpoints increase the diversity of the cyclists. Therefore, existing pedestrian detection approaches hardly achieve the acceptable detection performance.

Most previous methods have designed specific detectors using different features for each of these objects. The approach we claim here differs from these existing approaches in that we propose a single learning based detection framework to detect all the three important classes of objects. The proposed framework consists of a dense feature extractor and detectors of these three classes. Once the dense features have been extracted, these features are shared with all detectors. The advantage of using one common framework is that the detection speed is much faster, since all
dense features need only to be evaluated once in the testing phase. The proposed framework introduces spatially pooled features~\cite{paisitkriangkrai2014strengthening} as a part of aggregated channel features~\cite{DollarBMVC09ChnFtrs} to enhance the feature robustness to noises and image deformations. In order to further improve the generalization performance, we propose an object subcategorization method as a means of capturing intra-class variation of objects.

The remainder of this paper is organized as follows: we briefly review related works in Section~\ref{sec:Related_works}. The structure of the proposed detection framework will be discussed in Section~\ref{sec:Our_approach}. Experimental settings and results of all three applications are given in Section~\ref{sec:Experiments}. Section~\ref{sec:Conclusion} summaries this paper and points the direction of future work.

\section{Related works}
\label{sec:Related_works}

\subsection{Generic object detection}
\label{subsec:Generic object detection}

Object detection is a challenging but important application in the computer vision community. It has achieved successful outcomes in many practical applications such as face detection and pedestrian detection~\cite{viola2004robust,ahonen2004face,dalal2005histograms,wang2009hog}. Complete survey of object detection can be found in~\cite{viola2004robust,dalal2005histograms,felzenszwalb2010object,wang2013regionlets,girshick2013rich}. This section briefly reviews several generic object detection methods.

One classical object detector is the detection framework of Viola and Jones which uses a sliding-window search with cascaded classifiers to achieve accurate location and efficient classification~\cite{viola2004robust}. The other commonly used framework is using a linear support vector machine (SVM) classifier with histogram of oriented gradients (HOG), which has been applied successfully in pedestrian detection~\cite{dalal2005histograms}. These frameworks achieve excellent detection results on rigid object classes. However, for object classes with a large intra-class variation, their detection performance falls down dramatically~\cite{paisitkriangkrai2014strengthening}.

In order to deal with appearance variations in object detection, a deformable parts model (DPM) based method has been proposed in~\cite{felzenszwalb2010object}. This method relies on a variant of HOG features and window template matching, but explicitly models deformations using a latent SVM classifier. It has been applied successfully in many object detection applications~\cite{geiger2011joint,yebes14supervised,yan2013robust}. In addition to the DPM, visual subcategorization~\cite{divvala2012important} is another common approach to improve the generalization performance of detection model. It divides the entire object class into multiple subclasses such that objects with similar visual appearance are grouped together. A sub-detector is trained for each subclass and detection results from all sub-detectors are merged to generate the final results. Recently, a new detection framework which uses aggregated channel features (ACF) and a cascaded AdaBoost classifier has been proposed in~\cite{DollarPAMI14pyramids}. This framework uses exhaustive sliding-window search to detect objects at multi-scales. It has been adopted successfully for many practical applications~\cite{ohn2014fast,mathias2013traffic,paisitkriangkrai2014strengthening}.

\subsection{Traffic sign detection}
\label{subsec:Traffic sign detection}

Many traffic sign detectors have been proposed over the last decade with newly created challenging benchmarks. Interested reader should see~\cite{mogelmose2012vision} which provides a detailed analysis on the recent progress in the field of traffic sign detection. Most existing traffic sign detectors are appearance-based detectors. These detectors generally fall into one of four categories, namely, color-based approaches, shape-based approaches, texture-based approaches, and hybrid approaches.

Color-based approaches~\cite{de1997road,de2003traffic,kuo2007two} usually employ a two-stage strategy. First, segmentation is done by a thresholding operation in one specific color space. Subsequently, shape detection is implemented and is applied only to the segmented regions. Since RGB color space is very sensitive to illumination change, some approaches~\cite{fang2003road,maldonado2007road,kuo2007two} convert RGB space to HSI space which is insensitive to light change. Other approaches~\cite{stein1993hybrid,de1997road} implement segmentation in the normalized RGB space which is shown to outperform the HSI space~\cite{gomez2010goal}. Both HSI and normalized RGB space can alleviate the negative effect of illumination change, but still fail on some severe situations.

Shape-based approaches~\cite{houben2011single,loy2004fast,timofte2009multi} detect edges or corners from raw images using canny edge detector or its variants. Then, edges and corners will be connected to regular polygons or circles by using Hough-like voting scheme. These detectors are invariant to illumination change, but the memory and computational requirement is quite high for large images. In~\cite{de2003traffic}, the genetic algorithm is adopted to detect circles and is invariant to projective deformation, but the expensive computational requirement limits its application.

Texture-based approaches firstly extract hand-crafted features computed from texture of images, and then use these extracted features to train a classifier. Popular hand-crafted features include HOG, LBP, ACF, etc~\cite{dalal2005histograms,ahonen2004face,DollarPAMI14pyramids}. Some approaches~\cite{liang2013traffic,wang2013robust,pettersson2008histogram} use the HOG features with a SVM, others~\cite{mathias2013traffic} use the ACF features with an Adaboost classifier. Besides the above approaches, a convolutional neural network (CNN) has been adopted for traffic sign detection and achieved excellent results in~\cite{sermanet2011traffic}.

Hybrid approaches~\cite{gao2006recognition,prisacariu2010integrating} are a combination of the aforementioned approaches. Usually, the initial step is the segmentation to narrow the search space, which is same as the color-based approaches. Instead of only using edges features or texture-based features, these methods use them together to improve the detection performance.

One standard benchmark for traffic sign detection is the German traffic sign detection benchmark (GTSDB)~\cite{houben2013detection} which collects three important categories of road signs (prohibitory, danger, and mandatory) from various traffic scenes. All traffic signs have been fully annotated with the rectangular regions of interest (ROIs). Researchers can conveniently compare their work based on this benchmark.

\begin{figure*}[tbp]
\begin{center}
	\includegraphics[width=\linewidth]{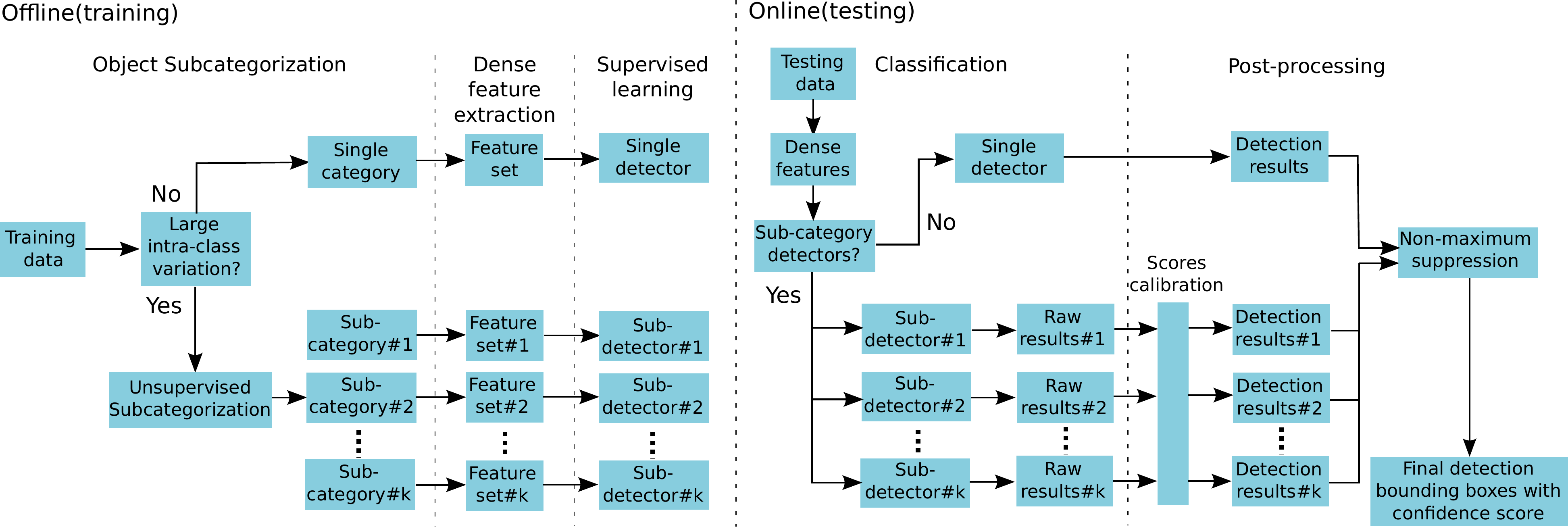}
\end{center}
\caption{Overview of the proposed detection framework. Left diagram is the training section and right diagram is the testing section.}
\label{fig:framework}
\end{figure*}

\subsection{Car detection}
\label{subsec:Car detection}

Many existing car detectors are vision-based detectors. Interested reader should see~\cite{sivaraman2003looking} which discusses different approaches for vehicle detection using mono, stereo, and other vision-sensors. We focus on vision-based car detectors using monocular information in this paper. These detectors can be divided into three categories: DPM-based approaches, subcategorization-based approaches and motion-based approaches.

DPM-based approaches are built on the deformable parts model (DPM)~\cite{felzenszwalb2010object} which has been successfully adopted in car detection~\cite{yebes14supervised}. In~\cite{geiger2011joint}, a variant of DPM discretizes the number of car orientations and each component of the mixture model corresponds to one orientation. The authors of~\cite{hejrati2012analyzing} train a variant of DPM to detect cars under severe occlusions and clutters. In~\cite{pepikj2013occlusion}, occlusion patterns are used as training data to train a DPM which can reason the relationships between cars and obstacles for detection. 

Visual subcategorization which learns subcategories within an object class is a common approach to improve the model generalization in car detection~\cite{divvala2012important}. It usually consists of two phases: feature extraction and clustering. Samples with similar visual features are grouped together by applying clustering algorithm on extracted feature space. Subcategorization-based methods are commonly used with DPM to detect cars from multiple viewpoints. In~\cite{kuo2009robust}, subcategories of cars corresponding to car orientation are learned by using locally linear embedding method with HOG features. In~\cite{ohn2014fast},  cars with similar viewpoints, occlusions, and truncation scenarios are grouped in the same subcategory by using a semi-supervised clustering method with ACF features.

Motion-based approaches often use appearance cues in monocular vision since monocular images do not provide any 3D and depth information. In~\cite{broggi2008lateral}, adaptive background model is used to detect cars based on motion that differentiated them from the background. The authors of~\cite{wang2005overtaking} propose an adaptive background model to model the area where overtaking cars tend to appear in the camera's field of view. Optical flow~\cite{martinez2008driving}, which is a popular tool in machine vision, has been used for monocular car detection. In~\cite{kyo1999robust}, a combination of optical flow and symmetry tracking is used for car detection. Optical flow is also used in conjunction with appearance-based techniques in~\cite{cui2010vehicle}.

The KITTI vision benchmark (KITTI)~\cite{geiger2013vision} is a novel challenging benchmark for the tasks of monocular, stereo, optical flow, visual odometry, and 3D object detection. The KITTI dataset provides a wide range of images from various traffic scenes with fully annotated objects. Objects in the KITTI dataset includes pedestrians, cyclists, and vehicles.

\subsection{Cyclist detection}
\label{subsec:Cyclist detection}

Many existing cyclist detection approaches~\cite{qui2003study,wang2006shape} use pedestrian detection techniques since pedestrians are very similar to cyclists along the road. In~\cite{qui2003study}, corner feature extraction, motion matching, and object classification are combined to detect pedestrians and cyclists simultaneously. In~\cite{wang2006shape}, a stereo vision based approach is proposed for pedestrian and cyclist detection. It uses the shape features and matching criterion of partial Hausdorff distance to detect pedestrians and cyclists. However, these approaches cannot distinguish cyclists from pedestrians. Besides the above approaches, the authors of~\cite{rogers2000counting} proposed a cyclist detector by using a fixed camera to detect two wheels of bicycle on road, but this approach is limited to detect crossing cyclists. Moreover, all above approaches are designed for traffic monitoring using fixed camera and cannot be used for on-road detection which aims at intelligent driving.

\section{Our approach}
\label{sec:Our_approach}
Despite several important techniques have been proposed on object detection, the conventional sliding-window based method of Viola and Jones~\cite{viola2004robust} is still the most successful and practical object detector. The VJ framework consists of two main components: a dense feature extractor and an AdaBoost classifier. In this paper, we build a common object detection framework for traffic scene perception based on the VJ framework, but our framework can employ a number of Adaboost classifiers to detect target objects of different classes. Apart from basic components of VJ framework, we propose an object subcategorization method to improve the generalization performance and employ spatially pooled features~\cite{paisitkriangkrai2014strengthening} to enhance the robustness and effectiveness.

Fig.~\ref{fig:framework} shows an overview of our framework. In the training phase, we firstly check the intra-class variation of the input object class with respect to object properties, e.g.\ size, orientation, aspect ratio, and occlusion. If the variation is considerable large, we apply the object subcategorization method to categorize the training data into multiple subcategories and train one sub-detector for each subcategory. Otherwise, we train a single detector for the entire training data. In the testing phase, raw detection results from all sub-detectors need to be calibrated before merging them together. Non-maximum suppression is used to eliminate redundant bounding boxes. If the framework employs detectors of different classes, detection results need to be carefully merged together.

\subsection{Object Subcategorization}
\label{subsec:Object Subcategorization}

For object classes with a large intra-class variation like cars, the appearance and shape of cars change significantly as the viewpoint changes. In order to deal with these variations that cannot be tackled by the conventional VJ framework, we present an object subcategorization method which aims to cluster the training data into visually homogeneous subcategories. The proposed subcategorization method applies an unsupervised clustering algorithm on one specific feature space of the training data to generate multiple subcategories. This method simplifies the original learning problem by dividing it into multiple sub-problems and improves model generalization performance.

\subsubsection{Visual Features}
A variety of hand-designed features can be used to perform the clustering algorithm, such as HOG and ACF~\cite{dalal2005histograms,DollarPAMI14pyramids}. HOG is successful at capturing the shapes of objects while does not consider color information. ACF combines both color information and gradient information, which is shown to outperform HOG~\cite{DollarBMVC09ChnFtrs}. In our experiments, a total of 10 channels of features are used for clustering: LUV color channels (3 channels), histogram of oriented gradients at 6 bins (6 channels), and normalized gradient magnitude (1 channel). To extract features from the training data, all positive samples are resized to the median object size.

\subsubsection{Geometrical Features}

Besides the visual features, geometrical information of objects can be extracted from traffic scenes using a variety of sensors and methods. In the KITTI dataset, objects in images from a velodyne laser scanner were annotated with 3D bounding boxes and 3D orientations. Ohn-Bar~\emph{et~al.}~\cite{ohn2015learning} proposed an analysis of different types of geometrical features, which showed that the geometrical features outperform the visual features for clustering, even the CNN features. We use the following set of geometrical features to represent the object instances in our experiments.

\textbf{3D orientation} The appearance and shape of objects change significantly as the viewpoint changes. We include the 3D orientation (relative orientation between the object and camera) in clustering, aiming at grouping objects with similar visual appearance together.

\textbf{Aspect-ratio} The aspect-ratio (width/height) of objects is strongly correlated with the geometry of objects being detected. We use this feature because learning models at different aspect-ratios significantly improve the generalization performance.

\textbf{Truncation level} The truncation level refers to the percentage of the object outside of the image boundaries. This feature strongly affects the appearance of objects.

\textbf{Occlusion index} Instead of using subtle occlusion patterns defined in~\cite{ohn2015learning}, we use an occlusion index to indicate whether an object is not occluded, partially occluded, largely occluded or an unknown situation. We simplify the occlusion patterns because some occlusion features cannot be defined for each object, such as occlusion level, related orientation, and relative 3D point. The above features can only be defined when the obstacle is an annotated object. However, in the KITTI dataset, many obstacles are unlabelled objects.

\subsubsection{Clustering}

A clustering algorithm is used to generate a predefined number of clusters on a specific feature space. Traditional clustering schemes, such as k-means or single linkage, suffer from the cluster degeneration which means that a few clusters claim most data samples~\cite{jain1999data}. The cluster degeneration problem can be alleviated by using spectral clustering. Spectral clustering followed by k-means often outperforms the traditional schemes. We implement the normalized spectral clustering using the algorithm proposed in~\cite{ng2002spectral}. The quality of clustering results is very sensitive to the predefined number of clusters. Unfortunately, how to determine the appropriate number of centroids is still an open question. We experimentally determine the number of clusters for each application.

\subsection{Feature extraction}
\label{subsec:Feature extraction}

The proposed framework introduces spatially pooled features~\cite{paisitkriangkrai2014strengthening} as a part of the aggregated channel features~\cite{DollarBMVC09ChnFtrs} and employs them as dense features in the training phase. All feature channels are aggregated in $4\times4$ blocks in order to produce fast pixel lookup features.

\subsubsection{Aggregated channel features (ACF)}

Given an input image $I$, a channel $C$ of $I$ is a feature map, where the output pixels are computed from corresponding pixels of the input image. Aggregated channel features are extracted from multiple image channels using pixel lookups method. Many image channels are available for extracting features. For example, a trivial channel of a grayscale image is the image itself. For a color image, each color channel can be used as a channel. Other channels can be computed using various transformations of $I$. In order to accelerate the speed of feature extraction, all transformations are required to be translational invariant. It means that the transformation need only to be evaluated once on the entire image rather than separately for each overlapping detection window.

ACF uses the same channel features as~\textbf{ChnFtrs}~\cite{DollarBMVC09ChnFtrs}: LUV color channels (3 channels), histogram of oriented gradients (6 channels), and normalized gradient magnitude (1 channel). ACF combines the richness and diversity of statistics from these channels, which is shown to outperform HOG~\cite{DollarPAMI14pyramids,DollarBMVC09ChnFtrs}. Prior to computing these 10 channels, we smooth the input image $I$ to suppress fine scale structures as well as noises.

\textbf{LUV color channels} LUV color space contains 3 channels, L channel describes the lightness of the object, U channel and V channel represent the chromaticity of the object. Compared to RGB space, LUV space is able to partially invariant to illumination change. So the proposed detector can work under different light conditions. Images can be converted to LUV space by using a specific transformation.

\textbf{Gradient magnitude channel} A normalized gradient magnitude is used to measure the edge strength. Gradient magnitude $M(x,y)$ at location $(x,y)$ is computed by $\sqrt{I_x^{2}+I_y^{2}}$, where $I_x$ and $I_y$ are first intensity derivatives along the $x$-axis and $y$-axis, respectively. Since the gradient magnitude is computed on 3 LUV channels independently, only the maximum response is used as the gradient magnitude channel.

\textbf{Gradient histogram channels} A histogram of oriented gradients is a weighted histogram where bin index is determined by gradient orientation and weighted by gradient magnitude~\cite{DollarBMVC09ChnFtrs}. The histogram of oriented gradients at location $(x,y)$ is computed by $M(x,y)\cdot\mathds{1}{[\Theta(x,y)=\theta]}$, where $\mathds{1}$ is the indicator function, $M(x,y)$ and $\Theta(x,y)$ are the gradient magnitude and discrete gradient orientation, respectively. ACF quantizes the orientation space to 6 orientations and compute one gradient histogram channel for each orientation.

\subsubsection{Spatially pooled features}

Spatial pooling is used to combine multiple visual descriptors obtained at nearby locations into a lower dimensional descriptor over the pooling region. We follow the work of~\cite{paisitkriangkrai2014strengthening} which is shown that pooling can enhance the robustness of two hand-crafted low-level features, covariance features~\cite{tuzel2006region} and LBP~\cite{ahonen2004face}.

\textbf{Covariance matrix} A covariance matrix is a positive semidefinite matrix which provides a measure of the relationship between multiple sets of variates. The diagonal elements of a covariance matrix represent the variance of each feature and non-diagonal elements represent the correlation between different features. In order to compute the covariance matrix, we use the following variates proposed in~\cite{paisitkriangkrai2014strengthening}:
\begin{displaymath}
[x, y, |I_x|, |I_y|, |I_{xx}|, |I_{yy}|, M, O_1, O_2]
\end{displaymath}
where $x$ and $y$ indicate the pixel location. $I_x$ and $I_y$ are first intensity derivatives along the horizontal-axis and vertical-axis respectively. Similarly, $I_{xx}$ and $I_{yy}$ are second intensity derivatives, respectively. $M$ is the gradient magnitude $\sqrt{I_x^{2}+I_y^{2}}$. $O_1$ is the edge orientation $\textrm{arctan}(|I_x|/|I_y|)$ and $O_2$ is an additional edge orientation in which,
\begin{displaymath}
 O_2 =
  \begin{cases}
   \textrm{atan2}(I_y, I_x) & \textrm{if atan2}(I_y, I_x) > 0, \\
   \textrm{atan2}(I_y, I_x) + \pi & \text{otherwise}.
  \end{cases}
\end{displaymath}
where the atan2 function is defined in terms of the arctan in the following:
\begin{displaymath}
\operatorname{atan2}(y, x) = \begin{cases}
\arctan\frac y x & \qquad x > 0 \\
\arctan\frac y x + \pi& \qquad y \ge 0 , x < 0 \\
\arctan\frac y x - \pi& \qquad y < 0 , x < 0 \\
+\frac{\pi}{2} & \qquad y > 0 , x = 0 \\
-\frac{\pi}{2} & \qquad y < 0 , x = 0 \\
\text{undefined} & \qquad y = 0, x = 0
\end{cases}
\end{displaymath}
The covariance descriptor of a region is a $9\times9$ covariance matrix which can be computed efficiently because the computational cost is independent of the size of the region. We also exclude the variance of pixel locations (x and y coordinates) and the correlation coefficient between pixel locations (x and y coordinates), since these features do not capture discriminative information. Due to the symmetry, each covariance descriptor finally contains $42$ different values. 

\begin{figure}[t]
\begin{center}
	\includegraphics[width=0.9\columnwidth]{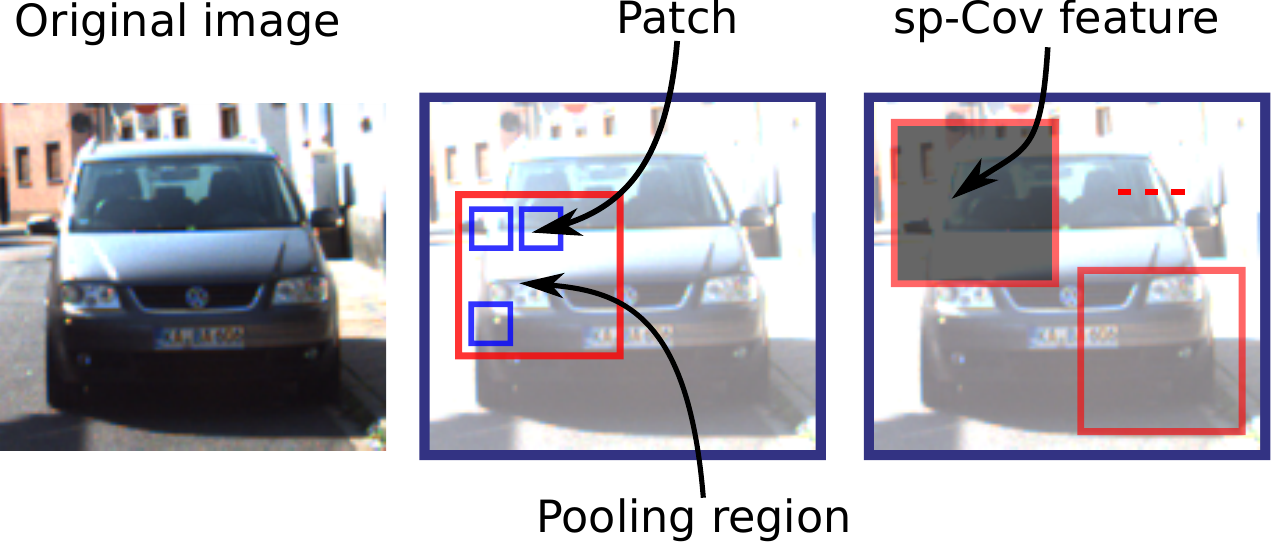}
\end{center}
\caption{Architecture of the spatially pooled covariance features.}
\label{fig:sp-Cov feature}
\end{figure}

\textbf{Spatially pooled covariance} The spatial invariance and robustness of the covariance descriptors can be improved by applying pooling method. There are two common pooling methods in this context: average pooling and max pooling. Max pooling is used in our framework as it has been shown to outperform average pooling in image classification~\cite{coates2011importance}. Max-pooling uses the maximum value of a pooling region to represent the pooled features in the region. It aims to retain the most salient information and discard irrelevant details and noises over the pooling region. The image window is divided into multiple dense patches (refer to Fig.~\ref{fig:sp-Cov feature}). Covariance features are computed over pixels within each patch. Then, we perform max pooling over a fixed-size pooling region and use the pooled features to represent the covariance features in the pooling region. In fact, multiple covariance matrices within each pooling region are summarized into a single matrix which has better invariance to image deformation and translation. The pooled features extracted from each pooling region is called the spatially pooled covariance (sp-Cov) features in~\cite{paisitkriangkrai2014strengthening}.

\textbf{Implementation} To expand the richness of our feature representation, we extract sp-Cov features using multi-scale patches with the following sizes: $4\times4$, $8\times8$ and $16\times16$ pixels. Each scale will generate an independent set of visual descriptors. In our experiments, the patch step-size is set to be 1 pixel, the pooling region is set to be $4\times4$ pixels, and the pooling spacing stride is set to be $4$ pixels.

\textbf{Local Binary Pattern (LBP)} LBP is a texture descriptor which uses a histogram to represent the binary code of each image patch~\cite{ahonen2004face}. The original LBP is generated by thresholding the $3\times3$-neighbourhood of each pixel with the value of centre pixel. All binary results are concatenated to form an 8-bit length binary sequence with $2^{8}$ different labels. The histogram of these 256 different labels can represent a texture descriptor. By following the work of~\cite{paisitkriangkrai2014strengthening}, we convert the input image from the RGB space to LUV space, and extract the uniform LBP~\cite{wang2009hog} from the luminance (L) channel. The uniform LBP, which is an extension of the original LBP, can better filter out noises.

\textbf{Spatially pooled LBP} Similar to the sp-Cov features, the image window is divided into multiple dense patches and LBP histogram is computed over pixels within each patch. In order to enhance the invariance to image deformation and translation, we perform max pooling over a fixed-size pooling region and use the pooled features to represent the LBP histogram in the pooling region. The pooled features extracted from each pooling region is called the spatially pooled LBP (sp-LBP) features in~\cite{paisitkriangkrai2014strengthening}.

\textbf{Implementation} To extract LBP, we apply the LBP operator on the 33-neighbourhood at each pixel. The LBP histogram is extracted from a $4\times4$ pixels patch. We extract the 58-dimension LBP histogram using a C-MEX implementation of~\cite{vedaldi2010vlfeat}. In our experiments, the patch step-size, the pooling region, and the pooling spacing stride are set to 1 pixel, $8\times8$ pixels, and 4 pixels, respectively. Instead of extracting LBP histograms from multi-scale patches, the sp-LBP and LBP are combined as channel features.

\subsection{Supervised learning}
\label{subsec:Supervised learning}

Once dense features have been extracted, we are in a position to train a classifier. Instead of training a standard AdaBoost, we use a shrinkage version of AdaBoost as the strong classifier and use decision trees as weak learners. To train the classifier, the procedure known as bootstrapping is applied, which collects the hard negative samples and re-trains the classifier. If the object subcategorization is applied to the training data, we train one classifier for each sub-detector. The pseudo code of the learning algorithm is presented in Algorithm~\ref{algorithm:AdaBoost}. 

\textbf{Shrinkage} The accuracy of AdaBoost can be further improved by applying a weighting coefficient known as shrinkage~\cite{hastie2005elements}. The shrinkage version of AdaBoost can be viewed as a form of regularization for boosting. At each iteration, the coefficient of weak learner is updated by
\begin{equation}
H_t(\vec{\boldsymbol{x}}) = H_{t-1}(\vec{\boldsymbol{x}})+ \nu \cdot w_{t}h_t(\vec{\boldsymbol{x}}).
\end{equation}
Here $h_t(\cdot)$ is a weak learner of AdaBoost at the $t$-th round and $w_{t}$ is the coefficient of the weak learner. $\nu{}\in(0, 1]$ is a learning rate which controls the trade-off between overall accuracy and training time. The smaller the value of $\nu$, the higher the overall accuracy as long as the number of weak learners is sufficiently large. Compared to the standard AdaBoost, shrinkage often produces better generalization performance~\cite{friedman2000additive}.

\textbf{Bootstrapping} To improve the performance of the learned classifier, we perform three bootstrapping iterations in addition to the original training phase. The initial training phase randomly sample negative samples from training images with positive regions cropped out, and further bootstrapping iterations add more hard negatives to the training set. The learning process consists of 4 training iterations with increasing numbers of weak learners and the final model consists of 2048 weak learners.

\def\ADot{ { $\bf \cdot$ } }%
\setcounter{AlgoLine}{1}
\begin{algorithm}[tbp]
\caption{\footnotesize Shrinkage version of AdaBoost}
\centering
{\footnotesize
   \begin{minipage}[]{0.94\linewidth}
    \vspace*{1mm}
    \KwIn{The training set $S = \{(\vec{x}_1,y_1), \cdots, (\vec{x}_i,y_i),\cdots, (\vec{x}_N,y_N)\}$,  $\vec{x}_i \subseteq \mathbb{R}^{n}$, $y_i \in \{-1, +1\}$, $i = 1,2,\cdots,N$.}
    \vspace*{1.5mm}
   { {\bf Initialize}:The weighted distribution $D$ of training set in 1st round, $D_1 = (w_{1,1}, \cdots, w_{1,i}, \cdots, w_{i,N}), w_{1,i} = 1/N$, $i = 1,2,\cdots,N$. }
    \vspace*{1.5mm}

   \For { $t = 1 \cdots T$  }
   {
    \ADot
	\vspace*{-1mm}
    	Train the weak learner $h_t(\cdot)$ using the weighted distribution $D_t$,
	\begin{displaymath}
	\vspace*{-1.5mm}
	h_t(\cdot): \mathbb{R}^{n} \rightarrow \{-1, +1\}
	\end{displaymath}

    \ADot
	\vspace*{-1.5mm}
        Compute the error rate $e_t$ of $h_t(\cdot)$ in traning set $S$.
	\begin{displaymath}
	\vspace*{-2mm}
	e_t = \sum_{i=1}^{N} w_{t,i} \cdot  \mathds{1}{(h_t(\vec{x}_i) \neq y_i)}
	\end{displaymath}

    \ADot
	\vspace*{0.5mm}
        Compute the coefficient $w_t$ of $h_t(\cdot)$ and update it by multiplying shrinkage parameter $\nu$.
    \vspace*{-1mm} 
	\begin{displaymath}
    	\vspace*{-1.5mm} 
	\begin{split}
	w_t &= \frac{1}{2}\log\frac{1-e_t}{e_t} \\
	a_t &= \nu \cdot w_t
	\end{split}
	\end{displaymath}
	
    \ADot
	\vspace*{-1mm}
        Update the weighted distribution of the training set
	\begin{displaymath}
	\vspace*{-1mm}
	\begin{split}
	D_{t+1} &= (w_{t+1,1}, \cdots, w_{t+1,i}, \cdots, w_{t+1,N}) \\
	w_{t+1,i} &= \frac{w_{t,i}}{Z_t}\exp{(-a_ty_ih_t(\vec{x}_i))}, i = 1, 2, \cdots, N
	\end{split}
	\end{displaymath}
	\vspace*{-1.5mm}
	where $Z_t$ is a normalization factor,
	\begin{displaymath}
	\vspace*{-1.5mm}
	Z_{t} = \sum_{i=1}^{N}w_{t,i}\exp{(-a_ty_ih_t(\vec{x}_i))}
	\end{displaymath}
   }

    \KwOut{Final classifier
	\vspace*{-2.5mm}
	\begin{displaymath}
	H(x) = \textrm{sign}\left(\sum_{t=1}^{T}a_th_t(\vec{\boldsymbol{x}})\right).
	\end{displaymath}}

\end{minipage}
}
\label{algorithm:AdaBoost}
\end{algorithm}

\subsection{Post-processing}
\label{subsec:Post-processing}

Raw detection results are generated by applying the trained detectors on test images, but these results often contain some noises and redundant information. To improve detection performance, some techniques are used to post-process raw detection results.

\subsubsection{Calibration of confidence scores}

If we have multiple sub-detectors and apply them on test data, detection results of each sub-detector are required to merge together to generate the integrated results. However, the classifier of each sub-detector is learned with different training data, confidence scores of raw detection results output by individual classifiers need to be calibrated appropriately to suppress noises before merging them together. We address this problem by transforming the output of each classifier by a sigmoid regression to generate comparable score distributions~\cite{platt1999probabilistic,lin2007note}. For sample $i$ in subcategory $k$, its confidence score is the output of the ensemble classifier which is defined as
\begin{equation}
s_i^k = \sum_{t=1}^{T}a_th_t(\vec{x}_i^k),
\end{equation}
its calibrated score is defined as
\begin{equation}
g_i^k = \frac{1}{1+\exp(A_k \cdot s_i^k+B_k)},
\end{equation}
where $A_k$, $B_k$ are the learned parameters for the $k$-th subcategory of the following regularized maximum likelihood problem:
\vspace*{-2mm}
\begin{equation}
\operatorname*{arg\,min}_{A_k,B_k}-\sum_{i=1}^{N_k}\left[t_i\log{g_i^k} + (1-t_i)\log{(1-g_i^k)}\right],
\label{eq:optimization1}
\vspace*{-2mm}
\end{equation}
\begin{equation}
 t_i =
  \begin{cases}
   \frac{N_{+}+1}{N_{+} + 2} & \textrm{if } y_i = +1 \\
   \frac{1}{N_{-} + 2}  & \textrm{if } y_i = -1
  \end{cases}
  \textrm{, } i = 1, \cdots, N_k.
\end{equation}
The $g_i^k$ in equation~\ref{eq:optimization1} can be cancelled by reformulation:
\vspace*{-2mm}
\begin{equation}
\begin{split}
\operatorname*{arg\,min}_{A_k,B_k}\sum_{i=1}^{N_k}\big[(t_i-1)(A_k\cdot s_i^k+B_k) + \\
\log{(1+\exp(A_k\cdot s_i^k+B_k))}\big].
\end{split}
\label{eq:optimization2}
\end{equation}
$N_k$ is the total number of training examples for the $k$-th subcategory-specific classifier, $N_{+}$ is the number of positive examples, and $N_{-}$ is the number of negative examples.

\subsubsection{Non-maximum suppression (NMS)}

NMS aims to suppress redundant overlaps among the raw detection results. When multiple bounding boxes overlap, NMS will eliminate the lower-scored detections and retain the highest-scored detection. Pascal overlap score~\cite{everingham2010pascal} is used to determine the overlap ratio $a_0$ between two bounding boxes. The overlap ratio $a_0$ is defined as
\begin{equation}
a_0 = \frac{\textrm{area}(B_{1}\cap{B_{2}})}{\textrm{area}(B_{1}\cup{B_{2}})},
\label{eq:nms}
\end{equation}
where $B_{1}$ and $B_{2}$ are two different bounding boxes. If the overlap ratio $a_0$ exceeds a predefined threshold, bounding box with the lower confidence score is discarded.

\subsubsection{Fusion of detection results}

The proposed framework can detect multiple objects using detectors or sub-detectors of different classes. Suppose we have detection results generated from different detectors, there are probably some redundant detections on the results since different detectors may generate some overlapped bounding boxes. NMS is usually used to delete redundant bounding boxes. However, NMS is not suitable for all cases. Assume that a car is occluded by a cyclist, both the car and the cyclist are detected in the results. If their overlap ratio exceeds the threshold, NMS will simply delete the lower-scored detection, and retain the higher-scored detection. One true positive detection is removed in this case.

To solve the above problem, we merge all detection results in two steps. In the first step, we merge detection results which belong to the same class using the NMS. It means that we apply NMS to bounding boxes generated by either a single detector of one class or multiple sub-detectors of one class exclusively. Objects of the same class are easily detected redundantly by multiple sub-detectors or a single detector at different scales. NMS is used to remove these redundant detections. In the second step, we merge all remaining detection results of different classes without using NMS to generate the final bounding boxes.

\section{Experiments}
\label{sec:Experiments}
\subsection{Traffic sign detection on GTSDB dataset}
\label{subsec:Traffic sign detection on GTSDB dataset}

In this section, we conduct an experiment on traffic sign detection and evaluate our detector on the German Traffic Sign Detection Benchmark (GTSDB)~\cite{houben2013detection}.

\subsubsection{Dataset}

\begin{figure}[htbp]
\begin{center}
\includegraphics[width=0.9\columnwidth]{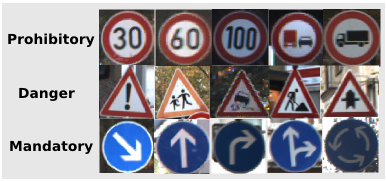}
\end{center}
\caption{Each row shows traffic signs in one of three categories (prohibitory, danger, mandatory).}
\label{fig:GTSDB signs}
\end{figure}

The GTSDB dataset contains 600 images for training and 300 images for testing. Images are captured from various scenes (highway, urban, rural) and various time slots (morning, afternoon, dusk,~\etc). The dataset contains more than 1000 traffic signs from different categories. Three main categories of traffic signs (Prohibitory, Danger, Mandatory) are selected as the target classes in the IJCNN 2013~\cite{houben2013detection} competition and in our experiments. The resolutions of traffic signs vary from $16\times16$ pixels to $128\times128$ pixels. Fig.~\ref{fig:GTSDB signs} shows traffic signs from three main categories on the GTSDB. 

\vspace*{-3mm}

\subsubsection{Evaluation criteria}

Pascal overlap score~\cite{everingham2010pascal} is used to find the best match between each predicted bounding box and each ground truth. The minimum overlap ratio $a_0$ is set to be 60\% on the GTSDB. Only the bounding box with the highest confidence score is counted as true positive if multiple bounding boxes satisfy the overlap criterion, the others are ignored. To compare the performance of different detectors, we follow the evaluation metric of the GTSDB which uses the area under the precision-recall curve (AUC) as a final score.

\subsubsection{Parameter selection}

To alleviate the effect of the illumination change, we apply the automatic color equalization algorithm (ACE)~\cite{getreuer2012automatic} to globally normalize all images. The resolution of the traffic sign model is set to $20\times20$ pixels and the dimension of model padding is set to $30\times30$ pixels. This border provides an additional amount of context that helps improve the detection performance~\cite{dalal2005histograms,DollarBMVC10FPDW}. Additionally, we increase the number of positive samples by adding jittered versions of the original samples, which significantly improves the detection performance. For prohibitory and danger signs, flipped versions are added to the training set. For mandatory signs, samples are randomly perturbed in translation ($[-2, 2]$ pixels), in scale ($[0.8, 1]$ ratio), in rotation ($[-5, 5]$ degrees), and flipping. We demonstrate the performance gain on the test set in table~\ref{tab:Jittered_performance}. Negative samples are collected from the GTSDB training images with the corresponding traffic sign regions cropped out.

\begin{table}[htbp]
\begin{center}
\begin{tabular}{c | c c c | c}
\hline
& & & & \\ [-2ex]
& Prohibitory & Danger & Mandatory & Avg. \\
\hline\hline
& & & &\\ [-2ex]
Original dataset & 98.76\% & 93.65\% & 86.86\% & 93.09\% \\

Jettered dataset & 100.00\% & 98.00\% & 97.57\% & 98.52\% \\
\hline
\end{tabular}
\end{center}
\caption{Performance (AUC) difference between training on original training set and jettered training set.}
\label{tab:Jittered_performance}
\end{table}

\subsubsection{Experimental design}

We investigate the experimental design of the proposed detector on traffic sign detection. Since traffic signs are divided into three subcategories, we train one sub-detector for each subcategory. We train all detectors on the GTSDB training set and evaluate them on the GTSDB test set. All experiments are carried out using combined features (ACF+sp-Cov+sp-LBP) as dense features, Adaboost with shrinkage value of 0.1 as the strong classifier, and depth3-decision trees as weak learners (if not specified otherwise).

\textbf{Shrinkage} We evaluate the performance of AdaBoost with 4 different shrinkage values from $\{0.05, 0.1, 0.2, 0.5\}$. We decrease the reject threshold of soft cascade by a factor of $\nu$ as coefficients of weak learners have been diminished by a factor of $\nu$. The area under precision-recall curve of different detectors are shown in Table~\ref{tab:GTSDB shrinkage}. We observe that applying a small shrinkage value often improves the detection performance and the best performance is achieved by setting $\nu=0.1$. However, without increasing the number of weak learners, setting the shrinkage value to be too small ($\nu=0.05$) can degrade the performance as the boosting cannot converge with a limited number of boosting iterations.

\begin{table}[htbp]
\begin{center}
\begin{tabular}{l | c c c | c}
\hline
& & & & \\ [-2ex]
Shrinkage & Prohibitory & Danger & Mandatory & Avg. \\
\hline\hline
& & & &\\ [-2ex]
$\nu = 0.5$ & 98.13\% & 95.28\% & 90.32\% & 94.58\% \\

$\nu = 0.2$ & 99.38\% & 96.80\% & 92.79\% & 96.32\% \\

$\nu = 0.1$ & \textbf{100.00\%} & \textbf{98.00\%} & \textbf{97.57\%} & \textbf{98.52\%} \\

$\nu = 0.05$ & 99.99\% & 97.81\% & 95.16\% & 97.63\% \\

$\nu = 0.05^{\ast}$ & 99.99\% & 98.00\% & 96.76\% & 98.25\% \\
\hline
\end{tabular}
\end{center}
\caption{Performance (AUC) of detectors with different shrinkage values. $\ast$~The model consists of 4096 weak learners while others consist of 2048 weak learners.}
\label{tab:GTSDB shrinkage}
\end{table}

\textbf{Depth of decision trees} We trained 4 different traffic sign detectors with decision trees of depth 1 to depth 4. Table~\ref{tab:GTSDB depth} shows the detection performance of different detectors. We observe that increase the depth of decision trees provides a performance gain, especially for the mandatory category. However, the depth-3 decision trees achieve better generalization performance and are faster to train than depth-4 decision trees.

\begin{table}[htbp]
\begin{center}
\begin{tabular}{c | c c c | c}
\hline
& & & & \\ [-2ex]
Depth & Prohibitory & Danger & Mandatory & Avg. \\
\hline\hline
& & & & \\ [-2ex]
depth-1 & 99.98\% & 97.41\% & 75.47\% & 90.95\% \\

depth-2 & 99.99\% & 97.98\% & 95.49\% & 97.82\% \\

depth-3 & \textbf{100.00\%} & \textbf{98.00\%} & 97.57\% & \textbf{98.52\%} \\

depth-4 & 99.99\% & 96.77\% & \textbf{98.10\%} & 98.29\% \\
\hline
\end{tabular}
\end{center}
\caption{Performance (AUC) of detectors with different depths of decision trees.}
\label{tab:GTSDB depth}
\end{table}

\textbf{Combination of features} To compare the discriminative power of different feature representations, we evaluate the performance of various feature combinations. The results are shown in Table~\ref{tab:GTSDB feature combination}. We observe that the combination of the sp-Cov features and LUV outperforms the ACF features and combining more features can further improve the detection performance. The best result is achieved by using the combination of all features (sp-Cov+sp-LBP+ACF).

\begin{table}[htbp]
\begin{center}
\begin{tabular}{l | c c c | c }
\hline
& & & & \\ [-2ex]
Feature combination & Prohibitory & Danger & Mandatory & Avg. \\ %
\hline\hline
& & & & \\ [-2ex]
ACF (LUV+O+M) & 98.72\% & 94.58\% & 92.65\% & 95.32\% \\%& 0.15s \\

sp-LBP+ACF & 99.99\% & 95.07\% & 96.12\% & 97.06\% \\%& 0.4s \\

sp-Cov+LUV & 99.30\% & 96.67\% & 95.56\% & 97.18\% \\%& 1.0s \\

sp-Cov+ACF & 98.73\% & 95.23\% & 95.61\% & 96.52\% \\%& 1.1s \\

sp-Cov+sp-LBP+ACF & \textbf{100.00\%} & \textbf{98.00\%} & \textbf{97.57\%} & \textbf{98.52\%} \\%& 1.5s \\
\hline
\end{tabular}
\end{center}
\caption{Performance (AUC) of detectors with various feature combinations.}
\label{tab:GTSDB feature combination}
\end{table}

\subsubsection{Comparison with state-of-the-art detectors}

Detection performance of various detectors on the GTSDB test set are shown in Table~\ref{tab:traffic sign detection results}. The proposed detector achieves the comparable results with state-of-the-art detectors despite its simplicity. These detectors~\cite{wang2013robust,mathias2013traffic} that offer better performance employ multi-scale models in detection. The authors of~\cite{wang2013robust} trained multiple subcategory-specific classifiers for each type of mandatory signs to achieve the best performance.

\begin{table}[htbp]
\begin{center}
\begin{tabular}{l|c|c|c|c}
\hline
& & & & \\ [-2ex]
Method & Prohibitory & Danger & Mandatory & Avg. \\
\hline\hline
& & & & \\ [-2ex]
Ours & \textbf{100.00\%} & 98.00\% & \textbf{97.57\%} & \textbf{98.52\%} \\

Wang~\emph{et~al.}~\cite{wang2013robust} & \textbf{100.00\%} & \textbf{99.91\%} & \textbf{100.00\%} & \textbf{99.97\%} \\

Mathias~\emph{et~al.}~\cite{mathias2013traffic} & \textbf{100.00\%} & \textbf{100.00\%} & \textbf{96.98\%} & \textbf{98.99\%} \\

BolognaCVLab~\cite{houben2013detection} & 99.98\% & \textbf{98.72\%} & 95.76\% & 98.15\% \\

Liang~\emph{et~al.}~\cite{liang2013traffic} & 100.00\% & 98.85\% & 92.00\% & 96.95\% \\

Timofte~\emph{et~al.}~\cite{timofte2009multi} & 61.12\% & 79.43\% & 72.60\% & 71.05\% \\

Viola-Jones~\cite{viola2004robust} & 90.81\% & 46.26\% & 44.87\% & 60.65\% \\
\hline
\end{tabular}
\end{center}
\caption{Detection performance (AUC) of various detectors on GTSDB test set with 60\% overlap ratio.}
\label{tab:traffic sign detection results}
\end{table}

\subsection{Car detection on UIUC dataset}
\label{subsec:Car detection on UIUC dataset}

Next, we conduct an experiment on car detection and compare detection performance of different detectors on the UIUC dataset~\cite{agarwal2004learning}. The UIUC dataset captures images of side views of cars with a resolution $40\times100$ pixels. The training set contains 550 positive samples and 500 negative samples. The test set is divided into two sets: 170 single-scale test images, containing 200 cars at roughly the same scale as in the training images, and 108 multi-scale test images, containing 139 cars at various scales.

We follow the evaluation protocol provided along with the UIUC dataset. A bounding box is counted as true positive if it lies within 25\% of the ground truth dimension in each direction. Only the bounding box with the highest confidence score is counted as true positive if multiple bounding boxes satisfy the criterion, the others are counted as false positives. In the dataset, three criteria are adopted to evaluate the performance: $F_1$-score, detection rate, and the number of false positives. $F_1$-score is the weighted harmonic mean of precision and recall.

The dimension of UIUC car model is set to $40\times100$ pixels without marginal padding as the car images are clipped to the same size. We expand the positive samples by flipping car images along the vertical axis. Since viewpoints of cars in the UIUC dataset are limited to side-views, we train a single detector without applying subcategorization and bootstrapping. Table~\ref{tab:UIUC car detection results} shows the results of different detectors on the multi-scale test subset. We observe that our detector achieves the best detection rate with slight more false positives on this dataset.

\begin{table}[htbp]
\begin{center}
\begin{tabular}{l|c|c|c}
\hline
& & & \\ [-2ex]
Method & F-Measure & Det. rate & No. false pos. \\
\hline\hline
& & & \\ [-2ex]
Ours & 98.6\% & 99.28\% & 3 \\

Pruning~\cite{paisitkriangkrai2014asymmetric} & 98.6\% & 97.8\% & 1 \\

AdaBoost~\cite{viola2004robust} & 98.6\% & 98.6\% & 2 \\

AdaBoost+LDA~\cite{wu2008fast} & 98.6\% & 97.8\% & 1 \\

CS-AdaBoost~\cite{masnadi2011cost} & 95.3\% & 95.5\% & 9 \\
\hline
\end{tabular}
\end{center}
\caption{Performance of various detectors on UIUC multi-scale test set.}
\label{tab:UIUC car detection results}
\end{table}

\begin{table}[tbp]
\begin{center}
\scalebox{0.9}
{
\begin{tabular}{|l|c|c|c|c|c|c|c|c|c|}
\hline
\multirow{2}{*}{}  & \multicolumn{2}{|c|}{Training} & \multicolumn{2}{|c|}{Testing} & \multicolumn{5}{|c|}{Properties}
\\ \cline{2-10}
& \begin{turn}{90}\# cars\end{turn} & \begin{turn}{90}\# images\end{turn} & \begin{turn}{90}\# cars\end{turn} & \begin{turn}{90}\# images\end{turn} & \begin{turn}{90}color\end{turn} & \begin{turn}{90}Annotations~~\end{turn} & \begin{turn}{90}multi-views\end{turn} & \begin{turn}{90}occ. labels\end{turn} & \begin{turn}{90}trunc. labels\end{turn}
\\ \hline
& & & & & & & & & \\ [-2ex]
UIUC Car & 550 & 1050 & 139 & 108 & & & & & \\
MIT Car & 516 & 516 & - & - & \checkmark & & & & \\
Street Parking & - & 881 & - & - & \checkmark & \checkmark & \checkmark & \checkmark & \\
Pascal VOC & 1250 & 713 & 1201 & 721 & \checkmark & \checkmark & \checkmark & \checkmark & \checkmark \\
\hline
& & & & & & & & & \\ [-2ex]
\textbf{KITTI Car} & 27k & 7481 & - & 7518 & \checkmark & \checkmark & \checkmark & \checkmark & \checkmark \\
\hline
\end{tabular}
}
\end{center}
\caption{Comparison of car datasets. The first four columns indicate the amount of training/testing data in each dataset. Note that KITTI dataset is two orders of magnitude larger than other existing datasets. The next five columns provide additional properties of each dataset.}
\label{tab:Car datasets comparison}
\end{table}

\subsection{Car detection on KITTI dataset}
\label{subsec:Car detection on KITTI dataset}

To further demonstrate the effectiveness and robustness of the proposed detector on car detection, we evaluate our detector on a more challenging object detection benchmark, KITTI dataset~\cite{geiger2013vision}

\subsubsection{Dataset}

The KITTI dataset is a recently proposed challenging dataset which consists of 7481 training images and 7518 test images, comprising more than 80 thousands of annotated objects in traffic scenes. Table~\ref{tab:Car datasets comparison} provides a summary of existing car datasets. We observe that the KITTI dataset provides a large number of cars with different sizes, viewpoints, occlusion patterns, and truncation scenarios. Due to the diversity of these objects, the dataset has three subsets (Easy, Moderate, Hard) with respect to the difficulty of object size, occlusion and truncation. Since the detection performance are ranked based on the moderately difficult results, we use the moderate subset as the training data in our experiments. The moderate subset contains 15710 cars, with the heights of the cars vary from 25 pixels to 270 pixels and the aspect ratios vary between 0.9 and 4.0. Since annotations of test data are not provided by the KITTI benchmark, we split the KITTI training images into training set (first 4000 images) and validation set (remaining 3481 images). 

\subsubsection{Evaluation criteria}

We follow the provided protocol for evaluation. Pascal overlap score is used to find the best match and the minimum overlap ratio $a_0$ is set to be 70\%. Only the bounding box with the highest confidence score is kept if multiple bounding boxes satisfy the overlap criterion, the others are counted as false positives. Instead of using AUC, average precision (AP)~\cite{everingham2010pascal} is used to evaluate the detection performance. The AP summaries the shape of the precision-recall curve, and is defined as the mean precision at a set of evenly spaced recall levels.

\subsubsection{Parameter selection}

We apply the proposed subcategorization method to categorize the training data into multiple subcategories. To find the model dimensions of each subcategory, we set the base height of each model to 52 pixels. From the base height, the width of each model can be obtained by taking the median aspect ratios of cars in the corresponding subcategory. Each model includes additional 4 pixels of margin on all sides. Using a model with suitable aspect ratio can significantly improves the detection performance due to better localization. We expand the positive training samples by randomly perturbing original car patches in translation ($[-2, 2]$ pixels), and in rotation ($[-2, 2]$ degrees). Negative samples are collected from the KITTI training images with vehicles regions cropped out.

\subsubsection{Experimental design}

We investigate the experimental design of the proposed detector on car detection. We train car detectors on the training set and evaluate them on the validation set. All experiments are carried out using ACF as dense features, Adaboost with shrinkage value of 0.1 as the strong classifier, depth4-decision trees as weak learners, and $K=25$ in the subcategorization method (if not specified otherwise). 

\begin{figure*}[tbp]
\begin{center}
	\includegraphics[width=0.9\linewidth]{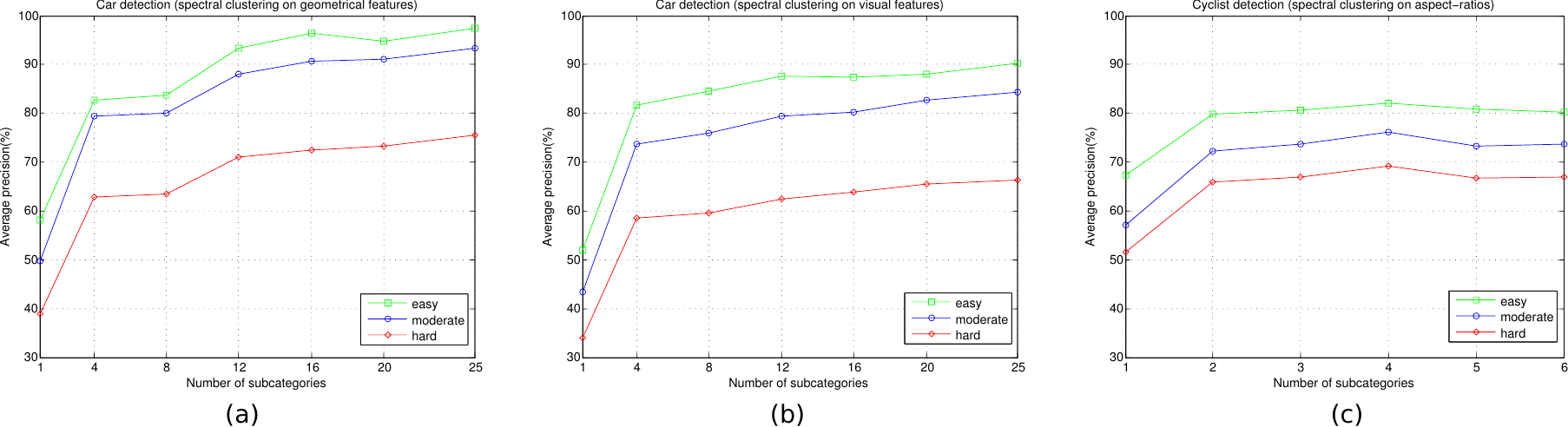}
\end{center}
\caption{(a) Average precision of spectral clustering + geometrical features with different number of subcategories on KITTI car validations set. (b) Average precision of spectral clustering + visual features with different number of subcategories on KITTI car validations set. (c) Average precision of spectral clustering + aspect-ratios with different number of subcategories on KITTI cyclist validations set.}
\label{fig:subcategorization_schemes}
\end{figure*}

\textbf{Number of subcategories} To investigate the effect of different numbers of clusters in our subcategorization method, we set the number from \{1, 4, 8, 12, 16, 20, 25\}. Fig.~\ref{fig:subcategorization_schemes}(a) and Fig.~\ref{fig:subcategorization_schemes}(b) shows the effect of increasing the number of subcategories on geometrical feature space and visual feature space, respectively. We observe that the geometrical features outperform the ACF features in the spectral clustering. We also observe that the detection performance improves as we increase the number of subcategories with $K=25$ provides the best performance.

\textbf{Depth of decision trees} We trained 4 different car detectors with decision trees of depth 2 to depth 5. Table~\ref{tab:Car depth} shows the average precisions of different detectors. We observe that depth-4 decision trees provide the best generalization performance.

\begin{table}[htbp]
\begin{center}
\begin{tabular}{c | c c c}
\hline
& & & \\ [-2ex]
Depth & Easy & Moderate & Hard \\
\hline\hline
& & & \\ [-2ex]
depth-2 & 96.38\% & 89.18\% & 70.87\% \\

depth-3 & 97.17\% & 91.21\% & 74.44\% \\

depth-4 & \textbf{97.41\%} & \textbf{93.37\%} & \textbf{75.60\%} \\

depth-5 & 96.67\% & 92.08\% & 72.77\% \\
\hline
\end{tabular}
\end{center}
\caption{Performance (AP) of detectors with different depths of decision trees.}
\label{tab:Car depth}
\end{table}

\textbf{Combination of features} We evaluate the performance of various feature combinations on car detection. The results are shown in table~\ref{tab:Car feature combination}. We observe that the detection performance improves as we add more features and the best performance is achieved by using the combination of all features (sp-Cov+sp-LBP+ACF). The combination of sp-LBP features and ACF features also achieves the similar performance and is five times faster than the combination of all features. We use the combination of sp-LBP features and ACF features as dense features in the testing since it gives a better trade-off between detection performance and runtime.

\begin{table}[htbp]
\begin{center}
\begin{tabular}{l | c c c | c}
\hline
& & & & \\ [-2ex]
Feature combination & Easy & Moderate & Hard & Runtime \\
\hline\hline
& & & & \\ [-2ex]
ACF (LUV+O+M) & 97.41\% & 93.37\% & 75.60\% & 0.5s \\

sp-LBP+ACF & 97.74\% & 94.38\% & 76.50\% & 1.5s \\

sp-Cov+LUV & 97.76\% & 93.68\% & 75.68\% & 6.8s \\

sp-Cov+ACF & 97.98\% & 93.48\% & 75.61\% & 6.8s \\

sp-Cov+sp-LBP+ACF & \textbf{98.42\%} & \textbf{94.55\%} & \textbf{76.66\%} & 7.5s \\
\hline
\end{tabular}
\end{center}
\caption{Performance (AP) of detectors with various feature combinations.}
\label{tab:Car feature combination}
\end{table}

\subsubsection{Comparison with state-of-the-art detectors}

Table~\ref{tab:KITTI car detection results} shows the performance comparison of state-of-the-art detectors on the KITTI testing set. Experimental results show that the proposed detector is of not only better performance than all DPM-based methods~\cite{pepikj2013occlusion,yebes14supervised,felzenszwalb2010object} but also less runtime. More significantly, our detector outperforms the SubCat~\cite{ohn2014fast} which employs a similar object subcategorization method and the Regionlets~\cite{wang2013regionlets,long2014accurate} which employs a similar pooling strategy. We conjecture that the additional performance gain is provided by the spatially pooled features.

\begin{table}[htbp]
\begin{center}
\begin{tabular}{l|c|c|c|c}
\hline
& & & & \\ [-2ex]
Method & Easy & Moderate & Hard & Runtime \\
\hline\hline
& & & & \\ [-2ex]
Ours & \textbf{87.19\%} & \textbf{77.40\%} & \textbf{60.60\%} & 1.5s \\

Regionlets~\cite{wang2013regionlets,long2014accurate} & 84.75\% & 76.54\% & 59.70\% & 1s \\

SubCat~\cite{ohn2014fast} & 81.94\% & 66.32\% & 51.10\% & 0.3s \\

AOG~\cite{li2014integrating} & 80.26\% & 67.03\% & 55.60\% & 3s \\

OC-DPM~\cite{pepikj2013occlusion} & 74.94\% & 65.95\% & 53.86\% & 10s \\

DPM-C8B1~\cite{yebes14supervised} & 74.33\% & 60.99\% & 47.16\% & 15s \\

MDPM-un-BB~\cite{felzenszwalb2010object} & 71.19\% & 62.16\% & 48.43\% & 60s \\

mBoW~\cite{behley2013laser} & 36.02\% & 23.76\% &18.44\% & 10s \\

\hline
\end{tabular}
\end{center}
\caption{Detection performance (AP) of various detectors on KITTI car test set with 70\% overlap ratio.}
\label{tab:KITTI car detection results}
\end{table}

\subsection{Cyclist detection on KITTI dataset}
\label{subsec:Cyclist detection on KITTI dataset}

In this section, we conduct an experiment on cyclist detection and evaluate our detector on the KITTI dataset.

\subsubsection{Dataset}

The KITTI dataset contains annotated cyclist objects which are captured from various traffic scenes. Similar to cars, cyclists are divided into three subsets (Easy, Moderate, Hard) and the moderate subset is used as the training data in our experiments. The moderate subset contains 1098 cyclists, with the heights of the cyclists vary from 25 pixels to 275 pixels and the aspect ratios vary between 0.3 and 1.5. 

\subsubsection{Evaluation criteria}

The KITTI cyclist detection uses the same evaluation protocol with the car detection expect that the minimum overlap ratio is relaxed to 50\%.

\subsubsection{Parameter selection}

The proposed subcategorization method is applied to cyclist detection. We define the dimensions of each cyclist model using the similar method in car detection. We set the base height of each model to 56 pixels, and the width of each model is derived from the median aspect ratios of cyclists in the corresponding subcategory. Each model includes additional 4 pixels of margin on all sides. We expand the positive training samples by randomly perturbing the original cyclists in translation ($[-2, 2]$ pixels), in rotation ($[-2, 2]$ degrees). Negative patches are collected from the KITTI training images with cyclist regions cropped out.

\subsubsection{Experimental design}

We investigate the experimental design of our detector on cyclist detection. We train cyclist detectors on the training set and evaluate them on the validation set. All experiments are carried out using ACF as dense features, Adaboost with shrinkage value of 0.1 as the strong classifier, depth4-decision trees as weak learners, and $K=4$ in the subcategorization method (if not specified otherwise).

\textbf{Number of subcategories} We set the number of clusters from \{1, 2, 3, 4, 5, 6\} in our subcategorization method. Since the minority of cyclists are occluded and truncated, clustering on all geometrical features leads to a cluster degeneration problem. We carefully select the aspect-ratios of cyclists as the feature space to avoid the above problem. Fig.~\ref{fig:subcategorization_schemes}(c) shows the effect of increasing the number of subcategories. We observe that the detection performance improves as we increase the number of subcategories when $K<=4$. Since the number of cyclists is much less than cars, the number of cyclists in each subcategory becomes very small when we have too many subcategories, which results in an imbalanced learning problem and degrading the detection performance.

\textbf{Depth of decision trees} We trained 4 cyclist detectors with decision trees of depth 2 to depth 5. Average precisions of different detectors are shown in Table~\ref{tab:Cyclist depth}. We observe that depth-4 decision trees offer the best generalization performance, as similar in the car detection.

\begin{table}[htbp]
\begin{center}
\begin{tabular}{c | c c c}
\hline
& & & \\ [-2ex]
Depth & Easy & Moderate & Hard \\
\hline\hline
& & & \\ [-2ex]
depth-2 & 80.92\% & 75.47\% & 69.46\% \\

depth-3 & 89.83\% & 82.67\% & 76.65\% \\

depth-4 & \textbf{92.15\%} & \textbf{86.18\%} & \textbf{79.28\%} \\

depth-5 & 90.98\% & 85.21\% & 78.26\% \\
\hline
\end{tabular}
\end{center}
\caption{Performance (AP) of detectors with different depths of decision trees.}
\label{tab:Cyclist depth}
\end{table}

\textbf{Combination of features} We evaluate the performance of various feature combinations on cyclist detection. The results are shown in Table~\ref{tab:Cyclist feature combination}. We observe that the best performance is achieved by using the combination of sp-LBP features and ACF features. The performance declines when we add the sp-Cov features as a part of aggregated channel features. The reason may be due to the lack of enough cyclist training data. We use the combination of sp-LBP features and ACF features as the dense features in the testing.

\begin{table}[htbp]
\begin{center}
\begin{tabular}{l | c c c | c}
\hline
& & & & \\ [-2ex]
Feature combination & Easy & Moderate & Hard & Runtime \\
\hline\hline
& & & & \\ [-2ex]
ACF (LUV+O+M) & 92.15\% & 86.18\% & 79.28\% & 0.2s \\

sp-LBP+ACF & \textbf{92.56\%} & \textbf{87.40\%} & \textbf{80.01\%} & 0.6s \\

sp-Cov+LUV & 85.48\% & 79.17\% & 72.20\% & 5.8s \\

sp-Cov+ACF & 85.16\% & 80.58\% & 73.64\% & 5.8s \\

sp-Cov+sp-LBP+ACF & 90.08\% & 83.80\% & 76.89\% & 6.1s \\
\hline
\end{tabular}
\end{center}
\caption{Performance (AP) of detectors with various feature combinations.}
\label{tab:Cyclist feature combination}
\end{table}

\subsubsection{Comparison with state-of-the-art detectors}

Table~\ref{tab:KITTI Cyclist detection results} shows the performance comparison with state-of-the-art approaches. As shown in Table~\ref{tab:KITTI Cyclist detection results}, our detector outperforms all other methods on the test set. Specifically, our detector outperforms the best DPM-based method DPM-VOC+VP~\cite{pepikj2015multi} on all the three subsets by 16.29\%, 14.95\%, and 12.35\%, respectively. Our detector also performs slightly better than the Regionlets~\cite{wang2013regionlets,long2014accurate}.

\begin{table}[htbp]
\begin{center}
\begin{tabular}{l|c|c|c|c}
\hline
& & & \\ [-2ex]
Method & Easy & Moderate & Hard & Runtime \\
\hline\hline
& & & \\ [-2ex]
Our method & \textbf{58.72\%} & \textbf{46.03\%} & \textbf{40.58\%} & 0.6s \\

Regionlets~\cite{wang2013regionlets,long2014accurate} & 56.96\% & 44.65\% & 39.05\% & 1s \\

MV-RGBD-RF & 52.97\% & 42.61\% & 37.42\% & 4s \\

DPM-VOC+VP~\cite{pepikj2015multi} & 42.43\% & 31.08\% & 28.23\% & 8s \\

LSVM-MDPM-us~\cite{felzenszwalb2010object} & 38.84\% & 29.88\% & 27.31\% & 10s \\

DPM-C8B1~\cite{yebes14supervised} & 43.49\% & 29.04\% & 26.20\% & 15s \\

mBoW~\cite{behley2013laser} & 28.00\% & 21.62\% & 20.93\% & 10s \\
\hline
\end{tabular}
\end{center}
\caption{Detection performance (AP) of various detectors on KITTI cyclist test set with 50\% overlap ratio.}
\label{tab:KITTI Cyclist detection results}
\end{table}

\subsection{An evaluation of the overall runtime}

We conduct an experiment on evaluation of the overall runtime with various feature combinations on the KITTI dataset. All experiments are carried out on a computer with an octa-core Intel Xeon 2.50GHz processor. The average runtime of each component of our detection framework can be seen in Table~\ref{tab:Overall_runtime}. For feature extraction, we observe that the ACF features can be extracted very quickly within 0.1s. When we add the sp-LBP features, the runtime increases moderately, but this features give an obvious performance gain in all three applications. When the sp-Cov features are employed, the runtime of feature extraction increases rapidly and dominates the total runtime of the system. For object detection, we observe that the car detector costs the most time in this framework since it has 25 sub-detectors. The traffic sign detector uses the least time since it has only 3 sub-detectors. We also observe that the runtime of detection increases as we add more features in this framework. According to observe the detection results of three applications, we conjecture that using a combination of ACF features and sp-LBP features can provide a better trade-off between detection performance and system runtime.

\begin{table}[htbp]
\begin{center}
\begin{tabular}{l@{\hskip 0.5mm} | @{\hskip 0.5mm}c@{\hskip 0.5mm} | @{\hskip 0.5mm}c@{\hskip 0.5mm} | @{\hskip 0.5mm}c@{\hskip 0.5mm} | @{\hskip 0.5mm}c@{\hskip 0.5mm} | @{\hskip 0.5mm}c}
\hline
& & & & & \\ [-2ex]
Feature combination & Feature & Cars(25) & Cyclists(4) & Signs(3) & Total \\
 & extraction & detection & detection & detection & Runtime \\
\hline\hline
& & & & & \\ [-2ex]
ACF (LUV+O+M) & 0.10s & 0.40s & 0.10s & 0.05s & 0.65s \\

sp-LBP+ACF & 0.35s & 1.20s & 0.30s & 0.10s & 1.95s \\

sp-Cov+ACF & 5.50s & 1.30s & 0.30s & 0.10s & 7.20s\\

sp-Cov+sp-LBP+ACF & 5.75s & 1.75s & 0.35s & 0.15s & 8.00s \\
\hline
\end{tabular}
\end{center}
\caption{An evaluation of the overall runtime of the proposed framework with various feature combinations.}
\label{tab:Overall_runtime}
\end{table}

\section{Conclusion}
\label{sec:Conclusion}

In this paper, we propose a common framework for detecting three important classes of objects in traffic scenes. The proposed framework introduces spatially pooled features as a part of the aggregated channel features to enhance the robustness and employs detectors of three important classes to detect target objects. The detection speed of the framework is fast since dense features need only to be evaluated once rather than individually for each detector. To overcome the weakness of the VJ framework for object classes with large intra-class variations, we propose an object subcategorization method to capture the variations and improves the generalization performance. We demonstrated that our detector achieves the competitive results with state-of-the-art detectors in traffic sign detection, car detection, and cyclist detection. Future work could include that contextual information can be used to facilitate object detection in traffic scenes and convolutional neural network can be used to generate more discriminative feature representations.\\

\textbf{Acknowledgements} This work was in part supported by Australia's Information and Communications Technology (ICT) Research Centre of Excellence.

{\small
\bibliographystyle{IEEE}
\bibliography{draft}
}

\begin{IEEEbiographynophoto}{Qichang Hu} is a PhD Candidature with the Australian Centre for Visual Technologies, University of Adelaide, Adelaide, SA, Australia.

He received the bachelor's degree in computer science from the University of Adelaide, Adelaide, SA, Australia in 2012. His research interests include deep learning, object detection, and machine learning.
\end{IEEEbiographynophoto}

\begin{IEEEbiographynophoto}{Sakrapee Paisitkriangkrai} is a Post-Doctoral Researcher with the Australian Centre for Visual Technologies, University of Adelaide, Adelaide, SA, Australia.

Dr. Paisitkriangkrai received the bachelor's degree in computer engineering, the master's degree in biomedical engineering, and the PhD degree from the University of New South Wales, Sydney, NSW, Australia, in 2003 and 2010, respectively. His research interests include pattern recognition, image processing, and machine learning.
\end{IEEEbiographynophoto}

\begin{IEEEbiographynophoto}{Chunhua Shen} is a Professor at School of Computer Science, the University of Adelaide. He was with the computer vision program at NICTA (National ICT Australia), Canberra Research Laboratory for about six years. His research interests are in the intersection of computer vision and statistical machine learning.

Dr. Shen studied at Nanjing University, Nanjing, China, and Australian National University, Canberra, ACT, Australia, and received the PhD degree from the University of Adelaide. From 2012 to 2016, he holds an Australian Research Council Future Fellowship.
\end{IEEEbiographynophoto}

\begin{IEEEbiographynophoto}{Anton van den Hengel} is a Professor at School of Computer Science, the University of Adelaide. He is also the Founding Director of the Australian Centre for Visual Technologies, Interdisciplinary Research Centre, University of Adelaide, Adelaide, SA, Australia, with a focus on innovation in the production and analysis of visual digital media.

Dr. van den Hengel received the bachelor's degree in mathematical science, the B.L. degree, the master's degree in computer science, and the PhD degree in computer vision from the University of Adelaide in 1991, 1993, 1994, and 2000, respectively.
\end{IEEEbiographynophoto}

\begin{IEEEbiographynophoto}{Fatih Porikli} is an IEEE Fellow and a Professor with the Research School of Engineering, Australian National University, Canberra, ACT, Australia. He is also acting as the Leader of the Computer Vision Group at NICTA, Sydney, NSW, Australia.

Dr. Porikli received the PhD degree from NYU, New York, NY, USA, in 2002. Previously he served as a Distinguished Research Scientist at Mitsubishi Electric Research Laboratories, Cambridge, MA, USA. He has contributed broadly to object detection, motion estimation, tracking, image-based representations, and video analytics. He is the coeditor of two books on Video Analytics for Business Intelligence and Handbook on Background Modeling and Foreground Detection for Video Surveillance. He is an Associate Editor of five journals including IEEE Signal Processing Magazine, SIAM Imaging Sciences, EURASIP Journal of Image \& Video Processing, Springer Journal on Machine Vision Applications, and Springer Journal on Real-time Image \& Video Processing. His publications won three best paper awards and he has received the R\&D100 Award in the Scientist of the Year category in 2006. He served as the General and Program Chair of several IEEE conferences in the past.
\end{IEEEbiographynophoto}

\end{document}